\documentclass{article}

\usepackage{PRIMEarxiv}

\usepackage[utf8]{inputenc} 
\usepackage[T1]{fontenc}    
\usepackage{hyperref}       
\usepackage{url}            
\usepackage{booktabs}       
\usepackage{amsfonts}       
\usepackage{nicefrac}       
\usepackage{microtype}      
\usepackage{lipsum}
\usepackage{fancyhdr}       
\usepackage{graphicx}       
\usepackage{enumitem}
\usepackage{algorithm}
\usepackage[table]{xcolor}
\usepackage{algpseudocode}
\newcolumntype{M}[1]{>{\centering\arraybackslash}m{#1}}
\graphicspath{{media/}}     

\title{Deep Image Composition Meets Image Forgery 
}

\author{
  Eren Tahir\\
  Yildiz Technical University \\
  Istanbul, Turkey\\
  \texttt{eren.tahir@std.yildiz.edu.tr} \\
    \And
  Mert Bal \\
  Yildiz Technical University \\
  Istanbul, Turkey\\
  \texttt{mertbal@yildiz.edu.tr} \\
}

\begin{document}
\maketitle

\begin{abstract}
Image forgery is a topic that has been studied for many years. Before the breakthrough of deep learning, forged images were detected using handcrafted features that did not require training. These traditional methods failed to perform satisfactorily even on datasets much worse in quality than real-life image manipulations. Advances in deep learning have impacted image forgery detection as much as they have impacted other areas of computer vision and have improved the state of the art. Deep learning models require large amounts of labeled data for training. In the case of image forgery, labeled data at the pixel level is a very important factor for the models to learn. None of the existing datasets have sufficient size, realism and pixel-level labeling at the same time. This is due to the high cost of producing and labeling quality images. It can take hours for an image editing expert to manipulate just one image. To bridge this gap, we automate data generation using image composition techniques that are very related to image forgery. Unlike other automated data generation frameworks, we use state of the art image composition deep learning models to generate spliced images close to the quality of real-life manipulations. Finally, we test the generated dataset on the SOTA image manipulation detection model and show that its prediction performance is lower compared to existing datasets, i.e. we produce realistic images that are more difficult to detect.
Dataset will be available at https://github.com/99eren99/DIS25k .
\end{abstract}

\keywords{image splicing \and image forgery \and image composition \and image manipulation detection}

\section{Introduction}
As technology has advanced, it has become easier to manipulate images and even ordinary people can produce difficult-to-detect manipulations through tools. Manipulated images are a serious threat that can affect people and communities financially and morally. To understand the scale of this threat, one can examine the funding and support for image manipulation detection research. 

Image editing tools have made and continue to make significant advances with the advent of artificial intelligence. Unfortunately, this progress has not been matched in image manipulation detection. Deep learning architectures and hardware advances have been equally effective for both image editing and manipulation detection. So what makes image manipulation detection lag behind image editing? The answer is big and high-quality data. Generative image models are trained with datasets of millions to billions of images, while image manipulation models are trained with datasets of a few thousand images. The difference is not only in quantity but also in quality. It is very costly to have a realistically manipulated image with ground truth prepared by an image editing expert. One solution to this problem is to automate image generation. So far, automated image generation frameworks can produce images that are much lower in quality than real-life manipulations.We use image composition techniques, which are directly related to image forgery, to improve the quality of the images produced.  

Image composition is the task of creating a new image by combining areas from different images. Although image composition is directly related to image manipulation, to the best of our knowledge, no previous work has addressed the two areas in terms of data generation like the framework we propose.

\subsection{Types of Image Forgery}
In real-life manipulations, a manipulator can use several of the methods classified below at the same time.
\begin{itemize}[leftmargin=*]
\item Image retouching: Less dangerous than the other methods when applied alone. It can be dangerous when used to hide other manipulations. It aims to alter some features of an image rather than changing the image drastically. Blurring, adding noise, morphological changes, contrast and brightness adjustments are some examples of this forgery.
\item Copy move: Copying a part of an image and pasting it into the same image, which can be detected by finding intra-image similarities or clipping artifacts.
\item Splicing: This is when a fragment from one image is cropped and pasted into another image. It can be detected by finding intra-image differences or clipping artifacts.
\item Removal inpainting: It is the process of filling a specified region in an image with inpainting methods. It can be detected by finding model-specific artifacts left by inpainting method used or intra-image differences.
\item Guided inpainting: The process of filling a defined region of an image with inpainting methods conditioned by text, images or both. It can be detected by the presence of intra-image differences or model-specific artifacts left by the inpainting method used. GANs and diffusion models are mostly used. Generative models, which are becoming increasingly sophisticated and begin to produce realistic results, pose a major threat.
\item Fully generated images: As known as text2image, image2image. This method has fewer intra-image inconsistencies than other methods. Forgery detection requires an inter-image approach rather than an intra-image approach. It can be detected by finding patterns that are not present in a normal image or model-specific artifacts. Recently, developers have been watermarking images to indicate that they are AI-generated, but there are effective methods to remove watermarks added to the image.
\end{itemize}

\subsection{Image Composition Substeps}
\begin{figure}
    \centering
    \includegraphics[width=1\linewidth]{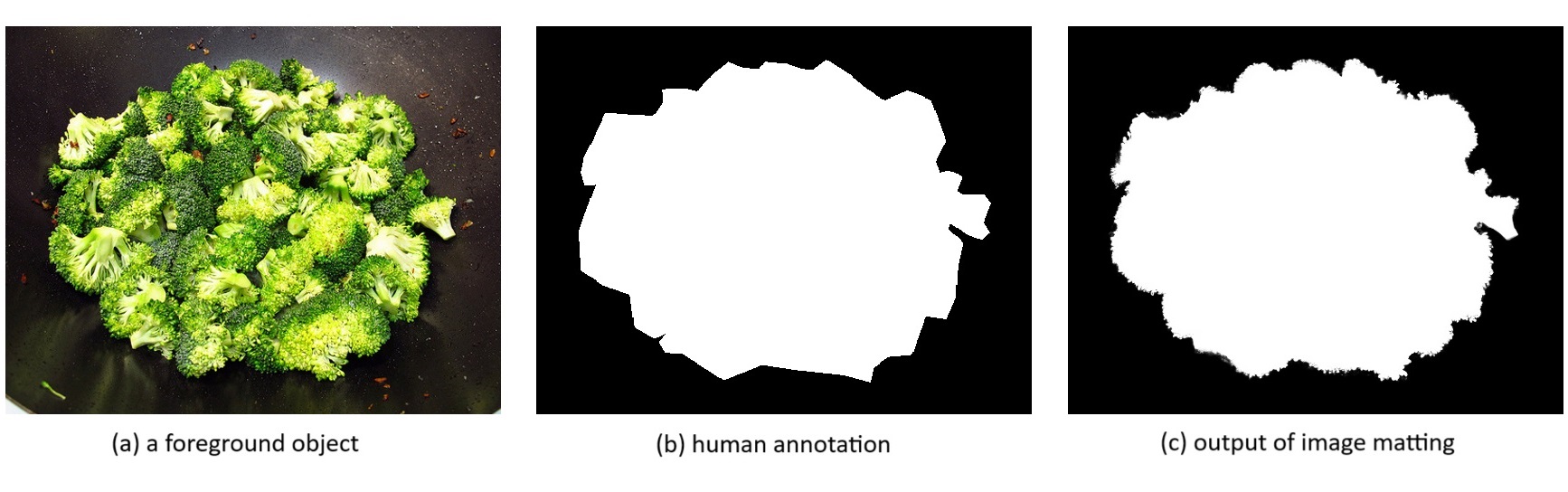}
    \caption{Illustration of image matting}
    \label{fig:1}
\end{figure}

\begin{figure}
    \centering
    \includegraphics[width=1\linewidth]{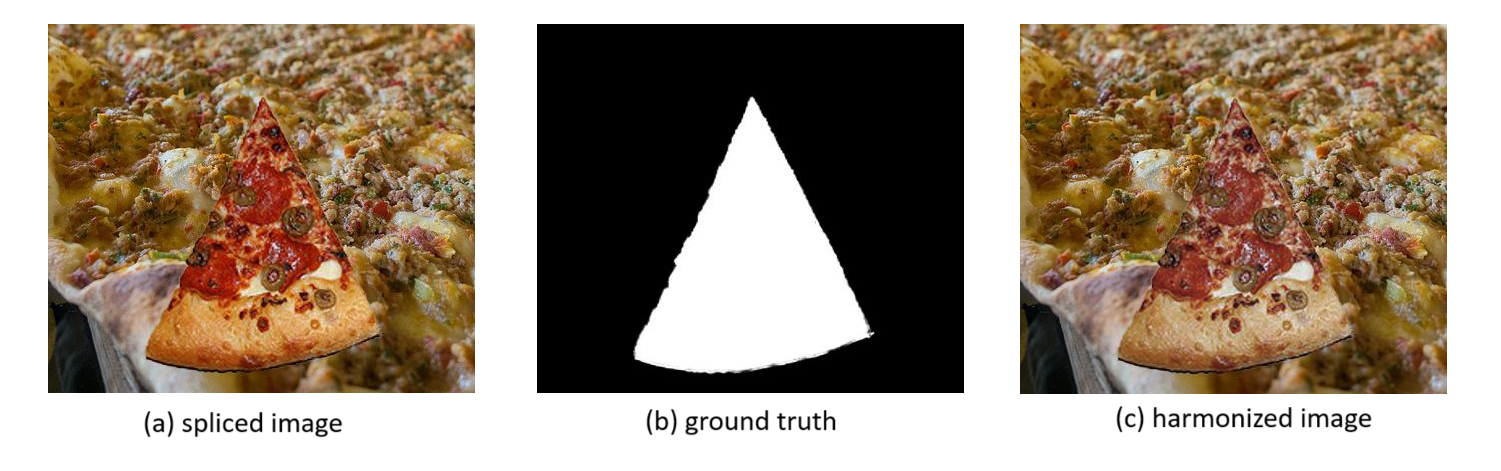}
    \caption{Illustration of image harmonization}
    \label{fig:2}
\end{figure}

\begin{itemize}[leftmargin=*]
\item Object placement: Determining the position of the foreground object to be pasted into the background image.
\item Image matting: Image matting is the process of separating the foreground object in the image at the pixel level. The result is a transparency value between 0-1 for each pixel. As the value approaches 1, the transparency of the pixel for the foreground object decreases "see Figure \ref{fig:1}".
\item Image blending: Image blending is the process of combining a foreground image with a background image. The goal of image blending is to create as seamless an image as possible while preserving the image content.
\item Image harmonization: In composite images, that is, images in which background and foreground images are combined, the foreground is altered so that it is indistinguishable from the background "see Figure \ref{fig:2}".
\end{itemize}

\section{Related Work}
"Image forgery detection confronts image composition"\cite{SCHETINGER2017152} is one of the few studies to address image composition and forgery together. At the time of publication in 2017, deep learning methods for image composition were not sufficiently developed and traditional methods were mostly used in the paper. In this study, 80 images were generated and qualitatively analyzed and it was concluded that the detection power with handcrafted features was very high. However, today, deep learning models are much more successful than traditional methods, but they cannot provide a reliable solution to real-life manipulations.

\begin{figure}
    \centering
    \includegraphics[width=1\linewidth]{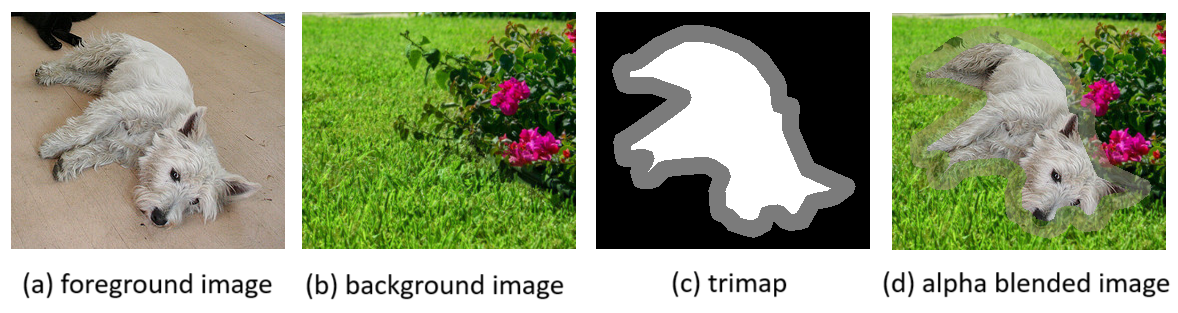}
    \caption{Artifacts of using alpha blending with trimap}
    \label{fig:3}
\end{figure}

DEFACTO dataset\cite{8903181}, to the best of our knowledge, one of two large-scale image splicing dataset with ground truths. It includes multiple types of manipulation techniques. 105,000 of its images are splicing images. While generating the splicing images, a trimap was created with morphological operations for the objects taken from the COCO\cite{Lin2014-ol} dataset. Afterwards, images were combined with alpha blending. Alpha blending is a weighted average of background and foreground images with a transparency mask
\[Composite Image =\alpha\circ(Foreground Image)+(1-\alpha)\circ(Background Image)\]

where \( \circ \) denotes Hadamard product and \( \alpha\) denotes transparency mask. When alpha blending and trimap are used directly, they create artifacts in gray regions and this makes it easier to detect manipulation "zoom Figure \ref{fig:3}". In our study, the gray areas in the trimap are enhanced with deep image matting, resulting in more difficult-to-detect manipulations. On the other hand, the other large-scale dataset SP COCO\cite{Kwon_2022} ,which has 200,000 images, generated by randomly pasting COCO objects on to images without preprocessing noisy COCO segmentation masks.

OPA dataset\cite{liu2021OPA} is a dataset for object placement. It was created by randomly pasting objects taken from the COCO dataset into appropriate background images and then labeling by humans whether the object positions are rational or not. The authors did publish metadata for their dataset including images' COCO IDs, object positions and object scales. Despite the generation purpose of this dataset, it could be used as a splicing dataset.

The Early Fusion image manipulation detection model\cite{triaridis2024exploring} is a deep learning model developed for image manipulation detection. It can perform both classification and localization. It shows the best performance on most of the available image forgery datasets\cite{paperswithcodePapersWithMMF}. It could be used for validating image forgery datasets' realism and quality.

\begin{table}
    \setlength{\tabcolsep}{6pt} 
    \renewcommand{\arraystretch}{1.35} 
    \setlength\arrayrulewidth{1pt} 
    \centering
    \begin{tabular}{|M{3cm}|M{1.5cm}|M{3cm}|M{2.5cm}|M{3cm}|M{0.8cm}|}
        \hline
        \rowcolor{lightgray}Name & Splicing Image Count & Preprocessing & Postprocessing & Method & Year\\
        \hline CAISA ITDE v1.0\cite{6625374}&921&Rotation, resizing, distortion&None&Manually created with Adobe Photoshop CS3&2013\\
        \hline CASIA ITDE v2.0\cite{6625374}&5123&Rotation, resizing, distortion&Blurring&Manually created with Adobe Photoshop CS3&2013\\
        \hline Columbia Gray\cite{articleColumbiaGray}&912&None&None&Manually created with Adobe Photoshop 6.0&2004\\
        \hline Columbia Color\cite{columbiaColor}&180&None&None&Manually created with Adobe Photoshop&2006\\
        \hline DSO-1\cite{6522874}&100&None&Colour and brightness adjustments&Manually created by researchers&2013\\
        \hline FantasticReality\cite{10.5555/3454287.3454307}&16000&?&?&?&2019\\
        \hline IMD2020 Real-Life Manipulated Images\cite{9096940}&fewer than 2010&?&?&Downloaded from internet&2020\\
        \hline In the wild\cite{Huh_2018_ECCV}&201&?&Resizing, blurring, jpeg compression&Downloaded from internet, manually annotated&2018\\
        \hline DEFACTO Splicing\cite{8903181}&105000&Trimap generation&None&Automated splicing with COCO dataset&2019\\
        \hline SP COCO\cite{Kwon_2022}&200000&Rotation, resizing& None&Automatic generation by randomly pasting COCO objects on to images&2022\\
        \hline Greatsplicing\cite{2310.10070}&5000&Scaling, rotation, skewing, distortion& Brightness, contrast, hue, saturation, color enhancements&Manually created with Adobe Photoshop&2023\\
        \hline AutoSplice\cite{10208982}&0&-&-&Even though its name is AutoSplice, inpainting forgery was implemented&2023\\
        \hline Ours&24964&Deep image matting&Deep image harmonization&Using OPA dataset and refining it with novel image composition techniques in an automated way&2024\\
        \hline Ours (Future Work)&100000&Both conventional and deep learning methods&Both conventional and deep learning methods&Using out-of-domain datasets and employing wide range of image composition techniques in an automated way&?\\
        \hline
    \end{tabular}
    \caption{Image splicing datasets that have ground truths}
    \label{tab:1}
\end{table}

\subsection{Shortcomings of previous datasets and critique on recent forgery detection studies}
None of the previously published splicing datasets have the characteristics of artifact diversity, high number of images, ground truth and realistic manipulations at the same time. The Table \ref{tab:1} shows how few images the datasets contain.

As explained in the introduction, the detection of manipulation types should be based on intra-image similarities, intra-image differences and inter-image relationships. A single model doing all these makes it difficult to reach a solution. Recent studies have tried to detect all types of manipulation with a single model. These models are especially unsuccessful in detecting fully generated images. A divide and conquer approach is more appropriate. This is one of the reasons why we generate a dataset specific to a single type(splicing) of manipulation. Other reasons are that generative methods are constantly evolving and producing a dataset for all types of manipulations requires a lot of human, time and hardware resources.

\section{Proposed Framework}
The novelty of the proposed framework is the automatic generation of image splicing dataset with image composition techniques. The aim of this work is not to make the best possible dataset, but to show what the proposed framework can do. It was done for POC purposes. 

To generate spliced images, one should find a foreground and a background image pair. After that, area from the foreground is extracted and pasted on the background image. To get rid of all of this overhead for now, the OPA dataset was used.Positions of the objects, COCO IDs of the objects and the background images provided in the OPA database were used during image generation process. Only the rationally labeled images are included in the work to generate realistic images.

In the image matting step, MatteFormer\cite{9879311} was used in order to enhance COCO objects' polygon style segmentation masks. MatteFormer is a transformer-based deep learning model developed for image matting. It has the 5th best score on Composition1k, one of the most popular image matting datasets \cite{paperswithcodePapersWith}. After enhancing segmentation masks, foreground objects and background images are combined via alpha blending.

Finally, in the image harmonization step, Harmonizer\cite{Harmonizer} was used. It is a model developed for image harmonization where whitebox filters and neural network is used together. It is a lightweight model developed for real-time video and image harmonization. Despite its lightweight structure, it has the 7th best score in iHarmony4, one of the most popular image harmonization datasets\cite{paperswithcodePapersWithHarmonizer}. Preserving the train test separation in the OPA dataset, 24964 splicing images were generated. Of the generated images, 3588 are test images and 21376 are train images. See Algorithm \ref{alg:1} for details.

\begin{figure}
    \includegraphics[width=1\textwidth]{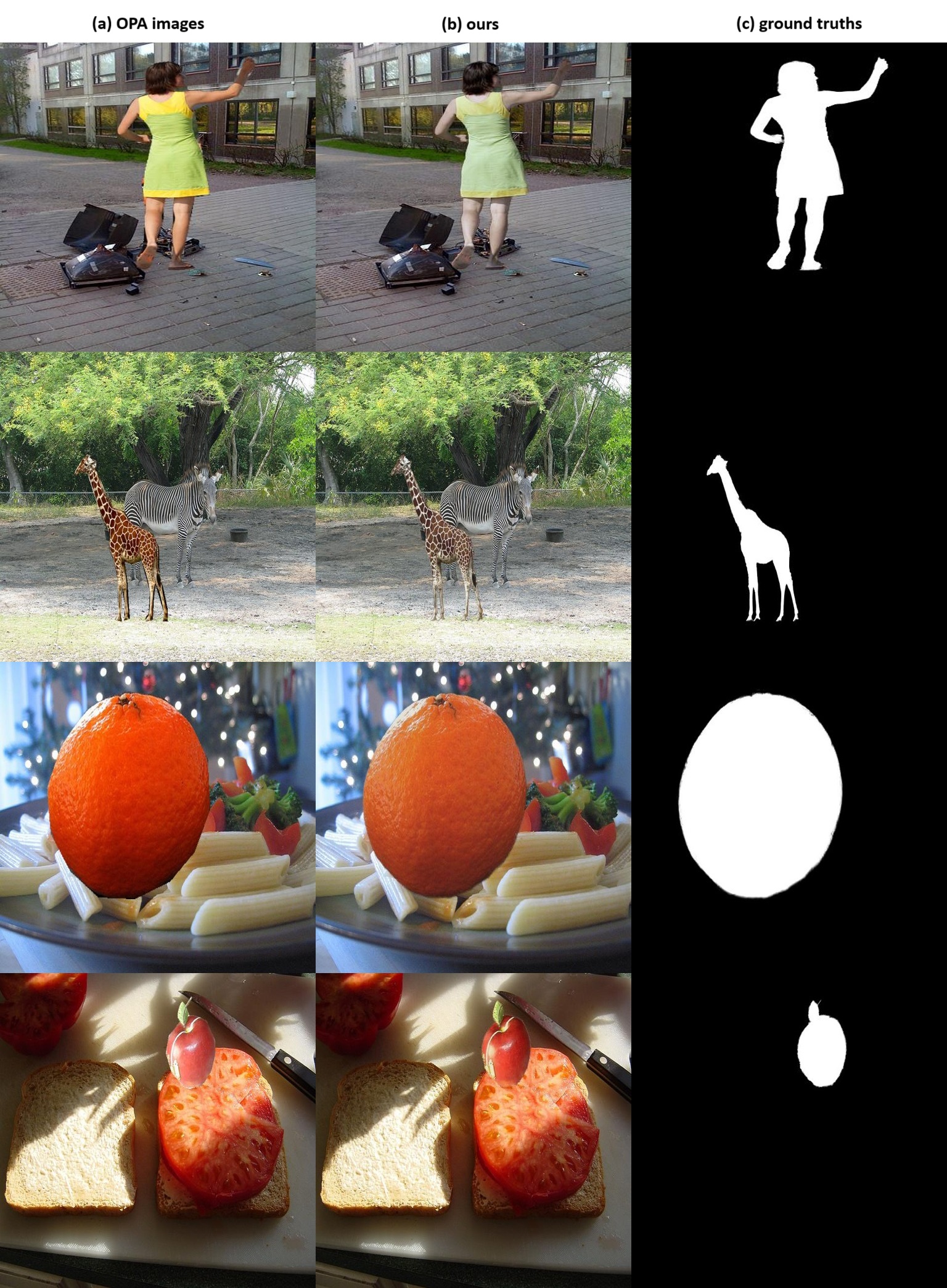}
    \caption{Comparison of proposed framework(zoom foreground objects' edges for details)}
    \label{fig:4}
\end{figure}

\begin{algorithm}
    \caption{Image generation}
    \label{alg:1}
    \begin{algorithmic}[1]
        \For {rationally labeled image in OPA}
            \State Get background image ID, foreground object COCO ID, object position, object height, object width from OPA metadata
            \State Download foreground object image with COCO API and generate segmentation mask with COCO metadata 
            \State Create a trimap for the foreground image by performing morphological operations on the segmentation mask
            \State Do deep image matting by giving the generated trimap and foreground object image to the MatteFormer model
            \State Combine the resulting mask, foreground object, background image with alpha blending. Create the ground truth mask
            \State Perform deep image harmonization by giving the combined image and ground truth to the Harmonizer model
        \EndFor
    \end{algorithmic} 
\end{algorithm}

\section{Evaluation}
In order to measure the quality of the dataset we produced, the Early Fusion image manipulation detection model was used to make predictions. Images were resized to 512x512 for convenience before making predictions. We taught it would not result in any accuracy drop because a very similar image manipulation detection study Trufor is not efected by image resizing \cite{guillaro2023trufor}. With resizing the SOTA detection model achieved 71.899\% classification success on 24964 images in our dataset. In addition, to test our hypothesis that image composition techniques help hide manipulation, we also had the model predict original OPA images. We found that the success rate was 78.38\% when image composition techniques were not applied to the same image composition. 

We did also check the accuracy of the SOTA model on other datasets. Despite the fact that image resizing (without changing image aspect ratio) data augmentation was employed during the training of the SOTA model, with resizing images to 512x512 resolution a drastic classification accuracy drop had been observed. One possible reason to this accuracy drop may be because of resizing by changing the image aspect ratio because the SOTA model has data augmentation without changing aspect ratio.

After this discovery, we did tested our dataset and corresponding OPA images without resizing. The detection model had 86.84\% accuracy on our dataset and  87.74\% on OPA images, in other words no significant accuracy change observed. Our guess on the reason for resizing without keeping aspect ratio affect our hypothesis test is the removal of compression and noise artifacts. Consequently our dataset, that has less spatial domain artifacts and color inconsistency, is harder to detect without noise and compression artifacts.

Compared to other datasets "see Table \ref{tab:2}", it achieved a lower success rate, which shows how high quality and difficult the images we produced were to detect. Unlike double compression at low quality level, adding noise and blurring the image, our proposed framework reduces the manipulation detection rate by 6.48\% without affecting the image quality.

\begin{figure}
    \centering
    \includegraphics[height=0.75\textwidth]{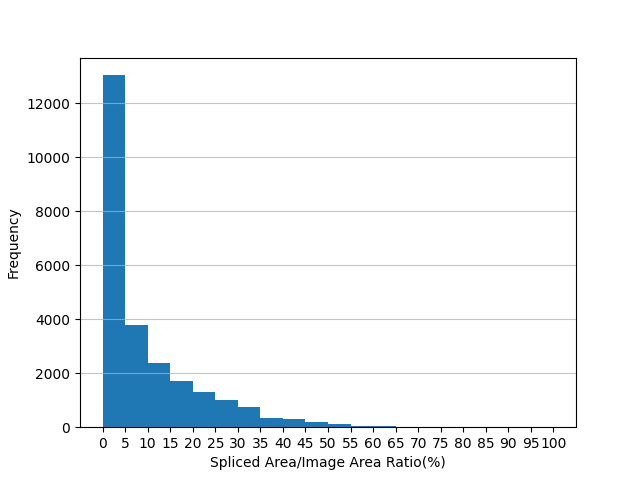}
    \caption{Histogram of manipulated area ratios}
    \label{fig:5}
\end{figure}

In splicing images, as a rule, the ratio of the area of the foreground object to the area of the composite image should not exceed 50\%. If it exceeds this, there will be confusion when identifying foreground and background regions. In the dataset we created, there are very few images where this ratio exceeds 50\% "see Figure \ref{fig:5}". Since this is a POC study, it can be neglected.

\begin{table}
    \renewcommand{\arraystretch}{1.05} 
    \centering
    \begin{tabular}{|M{3.8cm}|M{1.2cm}|M{1.2cm}|M{1cm}|M{0.9cm}|M{0.9cm}|M{2cm}|M{2cm}|}
        \hline
         & Columbia & Casiav1+ & DSO-1 & OPA & Ours & OPA(512x512) & Ours(512x512)\\
        \hline
        Classification Accuracy(\%)&96.2&84.5&93.5&87.74&86.94&78.38&71.899\\
        \hline
    \end{tabular}
    \caption{Forged image detection accuracy of the Early Fusion on image splicing datasets}
    \label{tab:2}
\end{table}

\section{Future Work}
To increase the coverage of the dataset by increasing artifact diversity, multiple deep learning models and traditional methods will be used in the matting, blending and harmonization stages. Some images will be generated without preprocessing and postprocessing.

In object placement stage, object placement network will be used to avoid being dependent on the OPA dataset.The low resolution objects in the COCO dataset both facilitate manipulation detection and adversely affect image matting. To avoid this, we will take objects from the salient object detection datasets and obtain high resolution objects. See Algorithm \ref{alg:2} for details.

\begin{algorithm}
    \caption{Image generation(future work)}
    \label{alg:2}
    \begin{algorithmic}[1]
        \State Pool=$\emptyset$
        \State Get objects and ground truths from salient object detection datasets
        \State Improve over 90\% of ground truths of foreground images with randomly selected image matting methods
        \State Classify foreground pictures by type in order to use this information while determining background
        \State Select random background images that match the classes of objects and append foreground\&background pair to the Pool
        \For {foreground\&background pair in Pool}
            \State Create rationality map via FastOPA\cite{niu2022fast} model
            \State Find the optimal location and object size on the generated rationality map with randomized search
            \State Perform blending with one of the following methods: alpha blending, laplacian blending, poisson blending
            \State Perform image harmonization with a 90\% probability with a randomly selected model from the harmonization methods
            \State Employ different kind of post processing attacks
        \EndFor
    \end{algorithmic} 
\end{algorithm}

\section{Conclusion}
Despite human, resource and hardware constraints, this study, which was carried out in a short time, demonstrates the success of the method. In future detailed studies, it is aimed to create a comprehensive dataset and close the gap of quality dataset in image splicing. We believe that working with higher resolution images and more advanced image composition methods will increase the rate of undetected manipulation.

Employing image composition methods for image forgery data generation should not be limited to image splicing. There is also a potential in copy move forgery for object placement to generate rational forgeries and image matting to reduce clipping artifacts.

\bibliographystyle{unsrt}  
\bibliography{references}

\end{document}